\documentclass{article}

\usepackage{microtype}
\usepackage{graphicx}
\usepackage{subfigure}
\usepackage{booktabs} %

\usepackage{hyperref}

\usepackage[accepted]{icml2025}

\usepackage{amsmath}
\usepackage{pifont}
\usepackage{amssymb}
\usepackage{mathtools}
\usepackage{amsthm}

\usepackage[capitalize,noabbrev]{cleveref}

\theoremstyle{plain}

\theoremstyle{definition}

\theoremstyle{remark}

\usepackage[textsize=tiny]{todonotes}

\icmltitlerunning{Scale-Distribution Decoupling: Enabling Stable and Effective Training of Large Language Models}

\begin{document}

\twocolumn[
\icmltitle{Scale-Distribution Decoupling: Enabling Stable and Effective \\Training of Large Language Models}

\icmlsetsymbol{equal}{*}

\begin{icmlauthorlist}
\icmlauthor{Ya Wang \textsuperscript{\dag}}{equal,yyy}
\icmlauthor{Zhijian Zhuo}{equal,yyy,comp}
\icmlauthor{Yutao Zeng}{equal,yyy}
\icmlauthor{Xun Zhou}{yyy}
\icmlauthor{Jian Yang}{}
\icmlauthor{Xiaoqing Li}{xxx}

\end{icmlauthorlist}

\icmlaffiliation{comp}{School of Mathematical Sciences, Peking University}
\icmlaffiliation{yyy}{Seed-Foundation-Model, ByteDance}
\icmlaffiliation{xxx}{Capital University of Economics and Business}
\icmlcorrespondingauthor{Xiaoqing Li}{xqli@cueb.edu.cn}

\icmlkeywords{Machine Learning, ICML}

\vskip 0.3in
]

\printAffiliationsAndNotice{\icmlEqualContribution \textsuperscript{\dag}Project lead} %

\begin{abstract}
Training stability is a persistent challenge in the pre-training of large language models (LLMs), particularly for architectures such as Post-Norm Transformers, which are prone to gradient explosion and dissipation. In this paper, we propose Scale-Distribution Decoupling (SDD), a novel approach that stabilizes training by explicitly decoupling the scale and distribution of the weight matrix in fully-connected layers. SDD applies a normalization mechanism to regulate activations and a learnable scaling vector to maintain well-conditioned gradients, effectively preventing \textbf{\textit{gradient explosion and dissipation}}. This separation improves optimization efficiency, particularly in deep networks, by ensuring stable gradient propagation. Experimental results demonstrate that our method stabilizes training across various LLM architectures and outperforms existing techniques in different normalization configurations. Furthermore, the proposed method is lightweight and compatible with existing frameworks, making it a practical solution for stabilizing LLM training. Code is available at \url{https://github.com/kaihemo/SDD}.
\end{abstract}

\section{Introduction} \label{introduction}

\begin{figure}[!ht]
\centering
\includegraphics[width=\linewidth]{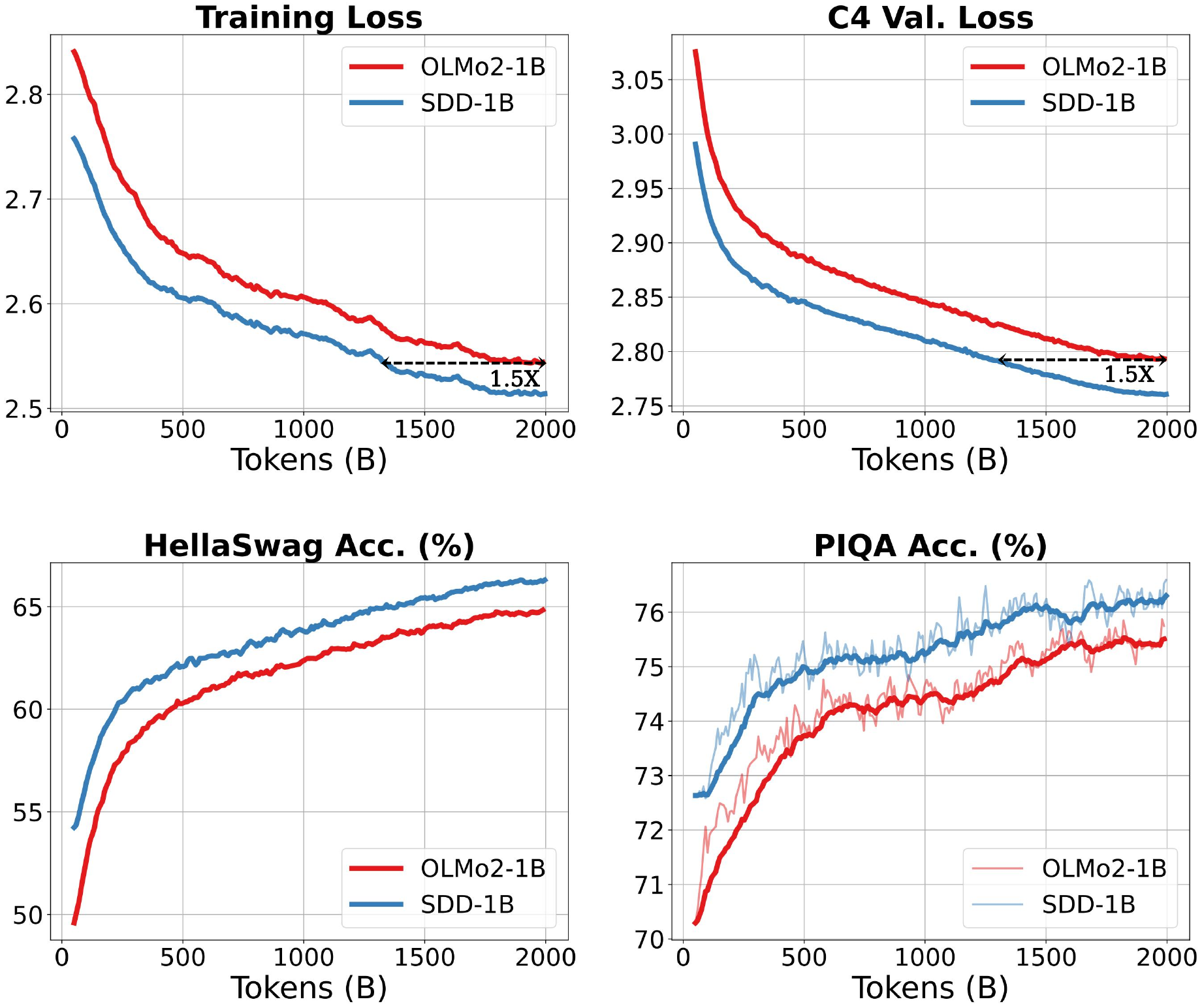}
\vspace{-18pt}
\caption{Training/validation loss with downstream performance on HellaSwag and PIQA for 1B dense models trained with 2T tokens: SDD-1B (Post-Norm) achieves superior convergence ($1.5\times$) and generalization over OLMo2-1B (Pre-Norm).
}
\vspace{-18pt}
\label{fig:overall evaluation results for dense model 2t}
\end{figure}

Large Language Models (LLMs) have demonstrated remarkable success in various natural language processing tasks \citep{li-etal-2024-knowcoder,zhu2024hyper,huang2025over}, fueled by advances in model architectures, large-scale datasets, and computational resources. However, the training stability of LLMs remains a critical challenge, especially as model size and complexity continue to grow. Instabilities during pre-training often lead to issues such as gradient explosion, vanishing gradients, or optimization stagnation, hindering the efficient and effective training of these models. Although Pre-Norm Transformer \citep{xiong2020layer,zhuo2025polynomial} architectures exhibit greater stability during training, they suffer from feature collapse \citep{wang2024deepnet,xie2023residual}, where representations across different layers become increasingly similar as depth increases. This phenomenon may contribute to the scaling bottleneck in large models. On the other hand, Post-Norm configurations remain significantly more difficult to train, exhibiting severe gradient explosion or vanishing issues, making stability in such settings a challenge in LLM research.

A fundamental source of these instabilities lies in the complexity of optimizing weight matrices in high-dimensional spaces. Specifically, the scale of weight parameters becomes challenging to regulate as the matrix grows in size, making convergence increasingly delicate. While existing strategies, such as sophisticated initialization schemes \citep{zhang2019improving} and normalization techniques \citep{ding2021cogview,xiong2020layer}, offer partial mitigation, they fail to resolve the core issue: the entanglement between the weight matrix's scale and distribution. This coupling induces suboptimal optimization dynamics, amplifying training instabilities, particularly in large-scale models where gradient propagation is susceptible to divergence or attenuation.

To tackle these challenges, we introduce \textbf{Scale-Distribution Decoupling (SDD)}, a novel approach that restructures fully-connected layers to explicitly separate the scale and distribution of weight matrices. In contrast to conventional formulations, SDD applies a normalization step to standardize activations, ensuring optimization focuses on learning the distribution rather than jointly optimizing both scale and distribution. Additionally, a learnable scaling vector is introduced to control the overall magnitude of activations, preventing gradient explosion and dissipation. This decoupling leads to more stable gradient propagation, enhancing both convergence efficiency and model robustness.

SDD is lightweight, requires minimal architectural modifications, and seamlessly integrates with a wide range of model configurations. Empirical evaluations demonstrate that SDD significantly improves training stability across various LLM architectures, including notoriously unstable Post-Norm Transformers. Furthermore, SDD accelerates convergence, improves generalization, and enables efficient large-scale pre-training, making it a practical and effective solution for stabilizing LLM training.

This work makes the following key contributions:
\begin{enumerate}
\item We introduce a novel design that explicitly decouples the scale and distribution of weight matrices, addressing a fundamental limitation in LLM optimization.
\item We empirically demonstrate that SDD stabilizes training across diverse LLM architectures, including both Pre-Norm and Post-Norm configurations, mitigating issues such as \emph{gradient explosion and dissipation}.
\item We provide empirical evidence showing that our method improves both convergence stability and training efficiency, making it highly applicable to large-scale pre-training tasks.
\end{enumerate}

The structure of this paper is organized as follows: Section \ref{sec:methodology} elaborates on the proposed methodology in detail. Section \ref{sc:theory} provides a theoretical analysis of the principles underpinning SDD. Experimental results and an in-depth analysis are presented in Section \ref{sec:experiments}. Section \ref{sec:related work} discusses related work on training stability and normalization techniques for large language models. Finally, Section \ref{sec:conclusion} concludes the paper with key insights and potential directions for future research.

\section{Scale-Distribution Decoupling} \label{sec:methodology}
\begin{figure}[!ht]
\centering
\includegraphics[width=\linewidth]{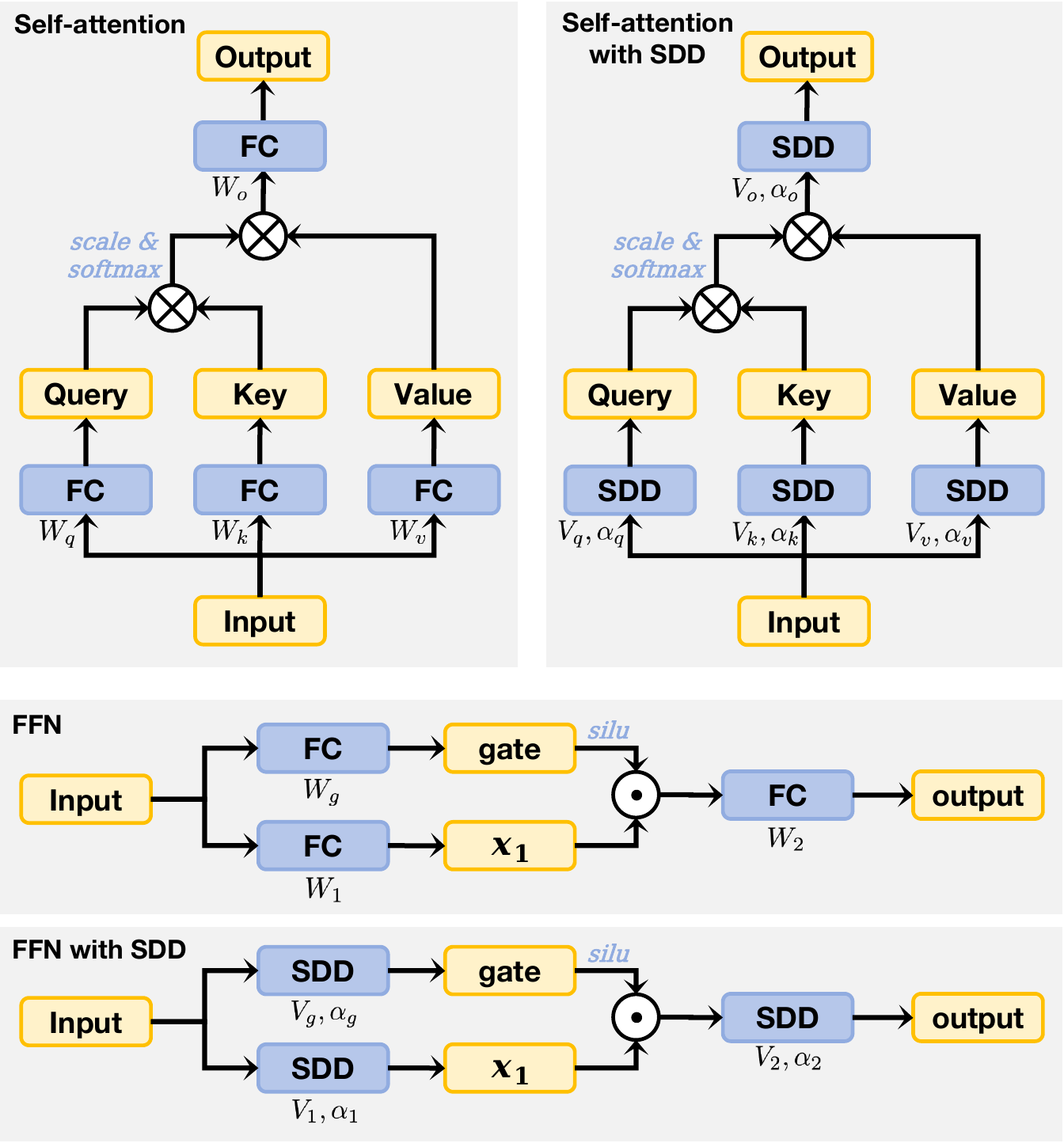}
\vspace{-18pt}
\caption{Comparison of vanilla and SDD-based Self-Attention /FFN Architectures. The top-left figure shows the standard self-attention module, while the top-right presents the self-attention module with SDD. Similarly, the middle figure depicts the standard feed-forward network (FFN), and the bottom shows the SDD-based FFN. In these figures, ``FC'' represents a fully-connected layer, and ``SDD'' denotes the SDD-based fully-connected layer, formulated as Eqn. \ref{equ:SDD}. Labels beneath ``FC'' and ``SDD'' indicate their learnable parameters. Notably, the additional parameter $\alpha$ in ``SDD'' is a one-dimensional vector, contributing negligible overhead.}
\vspace{-16pt}
\label{fig:architecture}
\end{figure}

\subsection{Motivation} \label{sec:motivation}
The training stability of large language models (LLMs) is frequently undermined by the challenges of optimizing high-dimensional weight matrices. Specifically, the scale of weight parameters has a profound impact on model outputs and gradient magnitudes but is inherently difficult to learn effectively. Existing techniques, such as advanced initialization schemes and normalization strategies, provide partial mitigation but fail to address a fundamental issue: the entanglement of the weight matrix's scale and distribution. This entanglement introduces unnecessary complexity to the optimization process, especially in Post-Norm Transformers, which are more susceptible to instability.

To address this issue, we propose \textbf{Scale-Distribution Decoupling (SDD)}, which disentangles the scale and distribution of weights in fully-connected layers. By isolating these two components, SDD not only simplifies the learning dynamics but also notably improves the training stability.

\subsection{Method}
In conventional fully-connected layers, the output is computed as $y = Wx$, where $W \in \mathbb{R}^{n \times n}$ represents the learnable weight matrix and $x \in \mathbb{R}^n$ is the input vector. The SDD formulation modifies this operation as follows:
\begin{equation} \label{equ:SDD}
y = \alpha \odot \mathrm{norm}(V x),
\end{equation} 
where $V \in \mathbb{R}^{n\times n}$ is a learnable weight matrix, $\odot$ denotes the element-wise multiplication. $\mathrm{norm}(\cdot)$ is a normalization function that removes scale information while preserving the distribution of $V x$ and $\mathrm{norm}(x) = \frac{x}{\|x\|}$ with $\|x\|=\sqrt{(x_1^2 + x_2^2 + \cdots + x_n^2)/n}$, following the normalization commonly used in Layer Normalization (LN) \citep{ba2016layer,wang2022anchor}. $\alpha$ is a learnable scaling vector to stabilize training during early stages (Figure~\ref{fig:architecture}). 

This reformulation separates the roles of the weight matrix: $\mathrm{norm}(Vx)$ captures the distributional characteristics, while $\alpha$ independently governs the scale. Such a decoupling has two key advantages. First, it simplifies optimization by disentangling scale and distribution, reducing complex parameter interactions that hinder learning. Second, normalization ensures bounded outputs, which inherently prevents gradient-related issues such as explosion or vanishing. These properties make SDD particularly effective for training deep and wide models, improving convergence and stability in challenging architectures.

SDD introduces minimal computational and memory overhead compared to standard fully-connected layers. The additional FLOPs for SDD are $6BSH$, where $B$ is the batch size, $S$ is the sequence length, and $H$ is the hidden size, accounting for only $3/H$ of the total model FLOPs. The parameter overhead is similarly negligible, adding just m parameters from the scaling vector  $\alpha$, contributing $1/H$ to the total parameter count. Given that $H > 1024$ in typical settings, both FLOPs and parameter overheads are negligible. Furthermore, SDD's additional memory cost can be effectively eliminated through gradient checkpointing, making it a lightweight yet effective modification.

\section{Theoretical Analysis} \label{sc:theory}
The SDD method is supported by a theoretical foundation that demonstrates its validity and advantages under common assumptions. To begin, we show that the proposed decoupling is equivalent to the standard fully-connected operation under Gaussian assumptions.

\subsection{Expressiveness of Standard and SDD-Based Layers}
Let  $x \in \mathbb{R}^n$ be sampled from a standard Gaussian distribution $\mathcal{N}(0, I)$, and each element of  $W \in \mathbb{R}^{n\times n}$  be i.i.d. Gaussian random variables with mean 0 and variance $\sigma^2/n$. For any fully-connected layer  $y = W x$, there exists an approximate representation  $y = \alpha \odot \mathrm{norm}(V x)$, where  $\alpha \in \mathbb{R}^n $ is a vector and  $V \in \mathbb{R}^{n\times n}$  is an matrix derived from  $W $. Conversely, any output of the form  $y = \alpha \odot \mathrm{norm}(V x)$  can be approximately represented in the form  $y = W x$.

Its proof, demonstrating the approximate expressiveness between standard and SDD-based layers, is provided in Appendix~\ref{appendix:proof of theorem equ}. \textit{The expectation symbol $\mathbb{E}$ is omitted for brevity.}

This equivalence encapsulates the fundamental principle of Scale-Distribution Decoupling (SDD): disentangling the scale and distribution of the weight matrix  $W$. SDD achieves this by introducing a learnable scaling vector  $\alpha$  to regulate magnitude, while  $\mathrm{norm}(V x)$  preserves the distributional structure of the transformed input. By explicitly decoupling these components, SDD streamlines optimization, obviating the need to simultaneously learn both scale and distribution. This separation enhances numerical stability, as  $\alpha$  facilitates precise control over output magnitudes, while normalization ensures a well-conditioned distribution. Furthermore, SDD exhibits strong adaptability, seamlessly accommodating both orthogonal and general weight matrices  $V$, making it a versatile and robust solution across diverse neural architectures.

\subsection{Gradient Analysis: Standard vs. SDD Layers}

The gradients with respect to  $\alpha$,  $V$, and  $x$  in the SDD-based formulation  $y = \alpha \odot \mathrm{norm}(V x)$  differ significantly from those in the standard fully-connected layer  $y = Wx$:
\begin{enumerate}
    \item The gradient with respect to $\alpha$  is well-conditioned and bounded, enabling faster and more stable optimization of the scale parameter.
    \item The gradient with respect to  $V$  is constrained by the normalization operation, ensuring bounded updates and avoiding gradient explosion or vanishing.
    \item The gradient norm with respect to  $x$  is moderated by the normalization operation, preventing gradient explosion or vanishing.
\end{enumerate}

\begin{proof}
For the standard fully-connected layer  $y = Wx$, the gradient with respect to $W$, which encodes both scale and distributional properties, is:
\begin{equation}
\frac{\partial \mathcal{L}}{\partial W} = \frac{\partial \mathcal{L}}{\partial y} \cdot x^\top,
\end{equation}
where  $\frac{\partial \mathcal{L}}{\partial y}$  is the backpropagated gradient. The magnitude of  $\frac{\partial \mathcal{L}}{\partial W}$  is highly sensitive to the initialization of both  $W$  and  $x$. Poorly scaled $W$  or  $x$  can lead to gradient explosion or vanishing, complicating optimization. In contrast, the SDD-based formulation  $y = \alpha \odot \mathrm{norm}(V x)$  decouples these components, leading to the following gradient properties:

\textit{\noindent{Gradient with Respect to $\alpha$:}}
The scale parameter $\alpha$, is explicitly learned in the SDD formulation, with its gradient given by:
\begin{equation}
\frac{\partial \mathcal{L}}{\partial \alpha} = \frac{\partial \mathcal{L}}{\partial y} \odot \mathrm{norm}(V x).
\end{equation}
Since  $\mathrm{norm}(V x)$  is bounded due to the normalization operation,  $\frac{\partial \mathcal{L}}{\partial \alpha}$  remains stable and well-conditioned throughout training. Unlike the standard formulation, where scale and distribution are entangled in  $W$, the decoupling in SDD allows $\alpha$ to be optimized independently. This results in consistently larger and more stable gradient updates for $\alpha$, enabling faster convergence of the scale parameter.

\textit{\noindent{Gradient with Respect to $V$:}}
The distributional characteristics of the input are controlled by $V$ in the SDD formulation. Given  $z = V x$, the gradient of the loss function  $\mathcal{L}$  with respect to  $V$  is expressed as:
\begin{equation}
    \frac{\partial \mathcal{L}}{\partial V} = \frac{\partial \mathcal{L}}{\partial y} \cdot \frac{\partial y}{\partial V}.
\end{equation}
Since  $y = \alpha \odot \mathrm{norm}(z)$, we have:
\begin{equation}
    \frac{\partial y}{\partial V} = \alpha \odot \frac{\partial \mathrm{norm}(z)}{\partial V}.
\end{equation}
    
The chain rule gives:
\begin{equation}
\frac{\partial \mathrm{norm}(z)}{\partial V} = \frac{\partial \mathrm{norm}(z)}{\partial z} \cdot \frac{\partial z}{\partial V}.
\end{equation}
Using the formula for the gradient of the normalized vector:
\begin{equation}
\frac{\partial \mathrm{norm}(z)}{\partial z} = \frac{1}{\|z\|} \left( I - \frac{z z^\top}{n\|z\|^2} \right),
\end{equation}
and $\frac{\partial z}{\partial V} = x^\top$. Substituting this into the gradient of  $\mathcal{L}$  with respect to  $V$:
\begin{equation}
\frac{\partial \mathcal{L}}{\partial V} = \frac{\alpha}{\|z\|} \odot \frac{\partial \mathcal{L}}{\partial y} \cdot \left( I - \frac{z z^\top}{n\|z\|^2} \right) \cdot x^\top.
\end{equation}
Next, assuming that  $V$  and  $x$  are i.i.d. with elements following a standard normal distribution  $\mathcal{N}(0, \sigma^2)$, we further simplify the expression. Let  $\|V\|_F$  denote the Frobenius norm of  $V$, which is defined as:
\begin{equation}
\|V\|_F = \sqrt{\sum_{i,j} V_{i,j}^2}.
\end{equation}
Incorporating this definition, the gradient becomes:
\begin{equation}
\frac{\partial \mathcal{L}}{\partial V} \approx \frac{\alpha}{\|V\|_F} \odot \frac{\partial \mathcal{L}}{\partial y} \cdot \left( I - \frac{z z^\top}{n\|z\|^2} \right) \cdot \frac{x^\top}{\|x\|}.
\end{equation}
A key observation is that  $\frac{\partial \mathcal{L}}{\partial y}$  remains stable across layers, with its magnitude exhibiting minimal fluctuations as it propagates through the network. This stability will be formally demonstrated in the subsequent gradient analysis with respect to  $x$. Consequently, the gradient norm of  $\frac{\partial \mathcal{L}}{\partial V}$  is primarily determined by  $\|V\|_F$, ensuring robustness during training. Furthermore, this stability enables precise control over  $\frac{\partial \mathcal{L}}{\partial V}$  via adjusting the standard deviation ($\text{std}$) of $V$. By simply initializing $V$  with small values, we can enhance convergence speed and improve overall training efficiency.

\textit{\noindent{Gradient with Respect to $x$:}} In the standard fully-connected layer, the gradient of the loss $\mathcal{L}$ toward the input  $x$  is:
\begin{equation}
    \frac{\partial \mathcal{L}}{\partial x} = W^\top \cdot \frac{\partial \mathcal{L}}{\partial y}.
\end{equation}
The gradient depends entirely on the transpose of the weight matrix  $W$  and the backpropagated gradient  $\frac{\partial \mathcal{L}}{\partial y}$. In this formulation, the gradient magnitude is sensitive to the scale and condition of $W$, meaning poorly scaled or ill-conditioned weight matrices can lead to gradient explosion or dissipation. Large singular values in $W$  amplify the gradient norm, resulting in unstable optimization due to gradient explosion, while small singular values reduce the gradient norm, leading to gradient dissipation and slowed convergence.

The SDD formulation  $y = \alpha \odot \mathrm{norm}(V x)$ incorporates a normalization step for  $V x$, fundamentally altering the gradient behavior. For the gradient with respect to  $x$ :
\begin{equation}
    \frac{\partial \mathcal{L}}{\partial x} \approx \frac{\alpha}{\|x\|} \odot \frac{\partial \mathcal{L}}{\partial y} \cdot \left( I - \frac{z z^\top}{n\|z\|^2} \right)\frac{V}{\|V\|_F},
\end{equation}
Due to the SDD network design, hidden embedding $x$ typically follows a standard normal distribution  $\mathcal{N}(0, 1)$. According to Theorem 3.1.1 \citep{vershynin2018high}, $\|x\|$ lies within a small neighborhood of 1, i.e., $\|x\|\approx 1$. For simplicity, we set $\|x\|=1$ by default.
Hence, the gradient norm becomes:
\begin{equation}
\|\frac{\partial \mathcal{L}}{\partial x}\| \approx \|\frac{\partial \mathcal{L}}{\partial y}\|.
\end{equation}
This equality implies that the gradient magnitude is preserved during backpropagation, neither exploding nor vanishing. The combination of normalization and initialization ensures that the network maintains stable gradients, regardless of the depth or dimensionality of the layers.
\end{proof}

SDD enhances training stability by disentangling the scale and distributional components of the weight matrix. By introducing normalization into all fully-connected layers, SDD ensures gradients remain bounded, mitigating gradient explosion and dissipation. The learnable scaling vector $\alpha$ independently controls the scale, while the normalized transformation $\mathrm{norm}(V x)$ isolates the distribution, improving the conditioning of $V$. These properties simplify optimization, enabling more robust and efficient training, especially in architectures prone to instability such as Post-Norm Transformers or high-dimensional layers. By addressing core challenges in large-scale neural network training, SDD provides a versatile and effective framework for stability and scalability.

\section{Experiment} \label{sec:experiments}
We evaluate SDD on both dense and MoE models, measuring training stability, convergence speed, and downstream performance. Our experiments include large-scale benchmarks, ablation studies, and robustness tests. Results show that SDD consistently improves training efficiency, mitigates instability, and outperforms existing normalization techniques across various architectures and tasks.

\subsection{Experimental Setup}

\textbf{Backbones.}
We evaluate SDD on two Transformer architectures: a 1B dense model and an MoE model with 588M active parameters (3.4B in total), both using Pre-Norm as the baseline. The dense model follows OLMo2 \citep{olmo20242} with 16 layers, $d_{model} = 2048$, 32 heads, and GQA (8 groups). The MoE model follows OLMoE \citep{muennighoff2024olmoeopenmixtureofexpertslanguage} with 32 layers, $d_{model} = 1024$, 16 heads, and 64 experts (8 active per token). Both models are trained from scratch for fair evaluation, with architectural details summarized in Table~\ref{table:model_architecture} and full configurations provided in Appendix~\ref{appendix: architecture}. All models are trained on the OLMoE Mix dataset \citep{muennighoff2024olmoeopenmixtureofexpertslanguage}. We compare SDD against Pre-Norm (baseline), Post-Norm \citep{vaswani2017attention}, and DeepNorm \citep{wang2024deepnet}.

\textbf{Training Setup.}
We train all models using the AdamW optimizer ($\beta_1=0.9, \beta_2=0.95$) on 4096-token sequences. Baseline models follow OLMo2 \citep{OLMo} and OLMoE \citep{muennighoff2024olmoeopenmixtureofexpertslanguage} initialization, combining truncated normal \citep{OLMo} and Megatron-Init \citep{shoeybi2019megatron}. In SDD, the parameter $\alpha$ is initialized as $1/\sqrt{\text{layers}}$ for the output mappings of the attention and feed-forward networks (FFNs), and as 1 for other projections. The remaining parameters are initialized using a normal distribution $\mathcal{N}(0, 1/\sqrt{2.5 \cdot d_{model}})$, ensuring that the initial outputs are aligned with the baselines. The dense model uses a learning rate of $3e^{-4}$ (decaying to $1.5e^{-5}$), while the MoE model starts at $4e^{-4}$, both following a cosine schedule. Training is conducted on 64 NVIDIA H800 80GB GPUs with a global batch size of 1024 and a micro-batch size of 4 per device, using next-token prediction loss (NLL). We also use gradient clipping (max norm 1.0) and BF16 mixed precision for stable and efficient training.

\textbf{Evaluation.}
We evaluate SDD across benchmarks covering reasoning, commonsense understanding, and question answering. Reasoning tasks include ARC-Easy, ARC-Challenge \citep{clark2018think}, PIQA \citep{bisk2020piqa}, and MMLU \citep{hendrycksmeasuring}. Commonsense understanding is assessed via HellaSwag \citep{zellers2019hellaswag}, Winogrande \citep{sakaguchi2021winogrande}, SocialIQA \citep{sap2019socialiqa}, and CSQA \citep{talmor2019commonsenseqa}. For question answering, we use SciQ \citep{welbl2017crowdsourcing}, CoQA \citep{reddy2019coqa}, BoolQ \citep{clark2019boolq}, COPA \citep{gordon2012semeval}, and OBQA \citep{mihaylov2018can}. Performance is measured via accuracy and loss using the LM Eval Harness framework \citep{eval-harness}.

\begin{figure}[!t]
\centering
\includegraphics[width=\columnwidth]{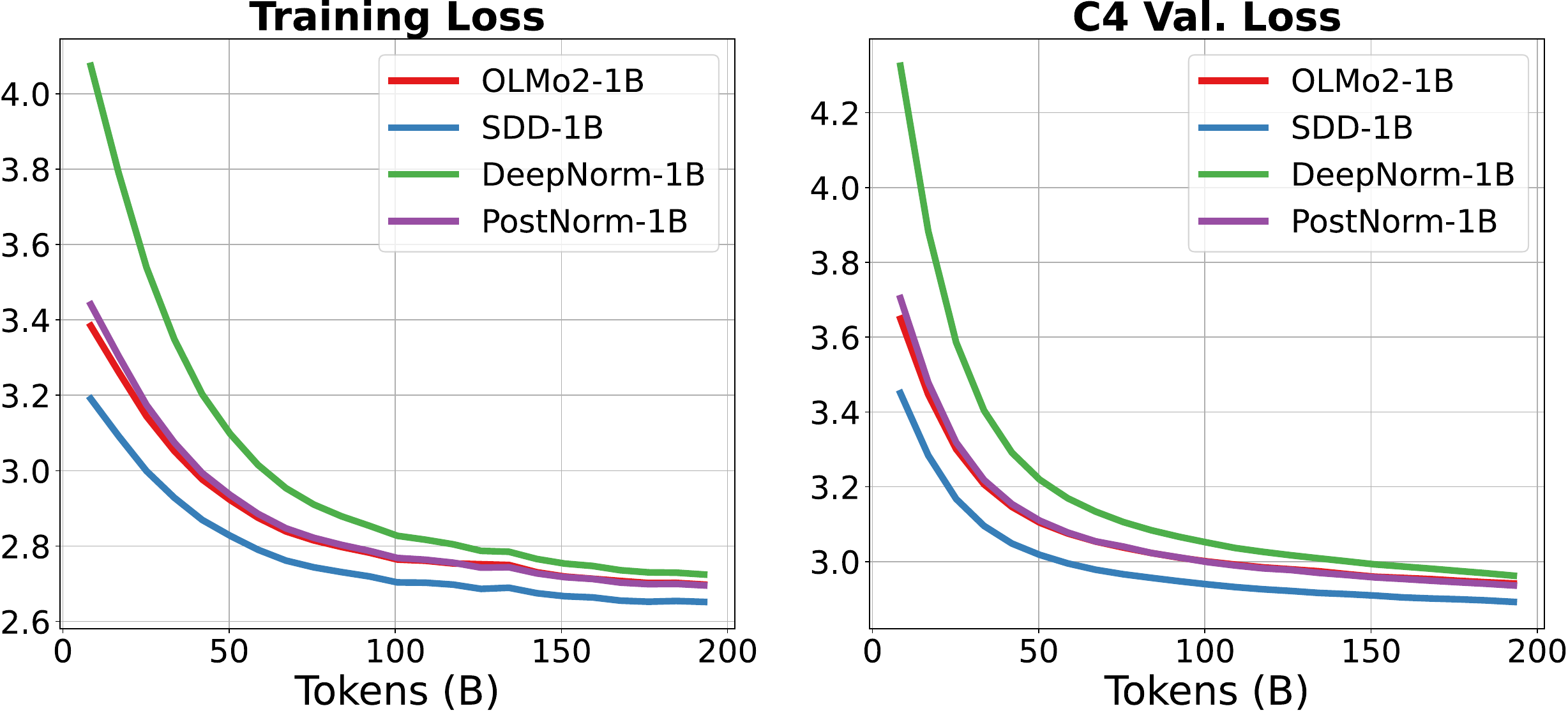}
\vspace{-18pt}
\caption{Training and validation loss on C4 for dense models trained with 200 billion tokens. A comparison of OLMo2-1B (Pre-Norm), DeepNorm-1B (Post-Norm), PostNorm-1B (Post-Norm), and SDD-1B (Post-Norm) highlights the superior convergence and stability of SDD-1B.
}
\vspace{-4pt}
\label{fig:overall training results for dense model}
\end{figure}

\subsection{Results on Dense Model}
We evaluate SDD on OLMo2-1B, a 1B-parameter dense model using Pre-Norm, comparing it to PostNorm-1B, DeepNorm-1B, and SDD-1B. PostNorm-1B and DeepNorm-1B are trained on 200B tokens, while OLMo2-1B and SDD-1B are trained on 2T tokens. For consistency, we report 200B token results here, with all evaluation metrics provided in Appendix~\ref{sec:additional resutls on dense model}, while full training dynamics for the 2T token runs are shown in Figure~\ref{fig:overall evaluation results for dense model 2t}. All models share identical hyperparameters, except DeepNorm-1B, which follows its official initialization scheme, ensuring a fair comparison.

\begin{figure}[!t]
\centering
\includegraphics[width=\linewidth]{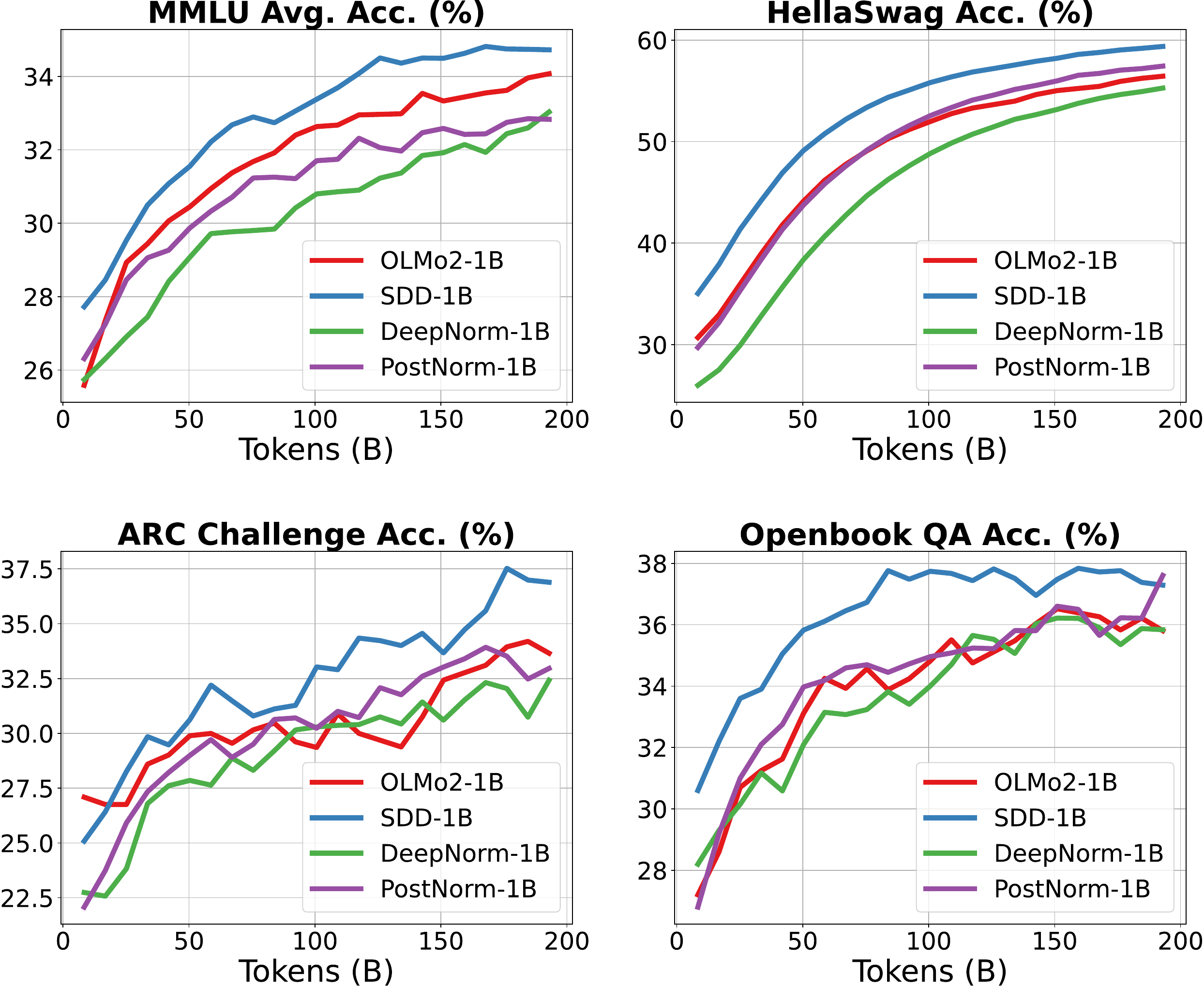}
\vspace{-18pt}
\caption{Downstream performance on MMLU, HellaSwag, ARC-Challenge, and OpenbookQA for dense models trained on 200B tokens. SDD-1B consistently outperforms others, showcasing superior generalization.
}
\vspace{-14pt}
\label{fig:overall evaluation results for dense model}
\end{figure}

\begin{table*}[!ht]
\caption{Performance comparison of the 1B dense models. This table compares training loss and downstream accuracy (\%). ``ARC-E'' and ``ARC-C'' denote ARC-Easy and ARC-Challenge. The best results are in bold, and ``Avg.'' represents average accuracy across tasks. SDD-1B achieves the best performance, demonstrating superior efficiency and generalization.}
\label{table: evaluation results on dense model}
\begin{center}
\setlength{\textwidth}{1pt}
\setlength{\tabcolsep}{4.0pt}
\begin{tabular}{l|c|cccccccccccccc}
\toprule
\textbf{Model}&\textbf{Loss $\downarrow$}&\textbf{MMLU}&\textbf{HellaSwag}&\textbf{ARC-C}&\textbf{ARC-E}&\textbf{Winogrande}&\textbf{Openbook QA}&\textbf{COPA}&\textbf{Avg. $\uparrow$} \\
\midrule
   OLMo2-1B  &      2.70     &       34.06        &       56.98        &         34.11          &       66.90       &        58.25        &        35.80         &     78.00     &  52.01  \\
 PostNorm-1B &      2.69     &       32.94        &       57.78        &         32.66          &       65.96       &        58.22        &        37.33         &     79.33     &  52.03  \\
 DeepNorm-1B &      2.72     &       33.06        &       55.73        &         31.77          &       65.09       &        55.99        &        35.67         &     79.67     &  51.00  \\
    SDD-1B   &\textbf{2.65}&\textbf{34.71}&\textbf{59.65}&\textbf{37.57}&\textbf{69.65}&\textbf{59.06}&\textbf{37.33}&\textbf{80.33}&\textbf{54.04} \\
\bottomrule
\end{tabular}
\end{center}
\vspace{-16pt}
\end{table*}

\textbf{Training Dynamics of 1B Dense Model.} Figure~\ref{fig:overall training results for dense model} shows the training and validation loss on C4 for 1B dense models trained with 200B tokens. Among OLMo2-1B (Pre-Norm), PostNorm-1B, DeepNorm-1B (both Post-Norm), and SDD-1B (Post-Norm), SDD-1B converges faster and reaches the lowest loss. It achieves 2.65, outperforming OLMo2-1B (2.70), PostNorm-1B (2.69), and DeepNorm-1B (2.72), demonstrating superior stability and efficiency. These results highlight SDD's ability to improve optimization by decoupling scale and distribution.

\begin{figure}[!ht]
\centering
\includegraphics[width=\columnwidth]{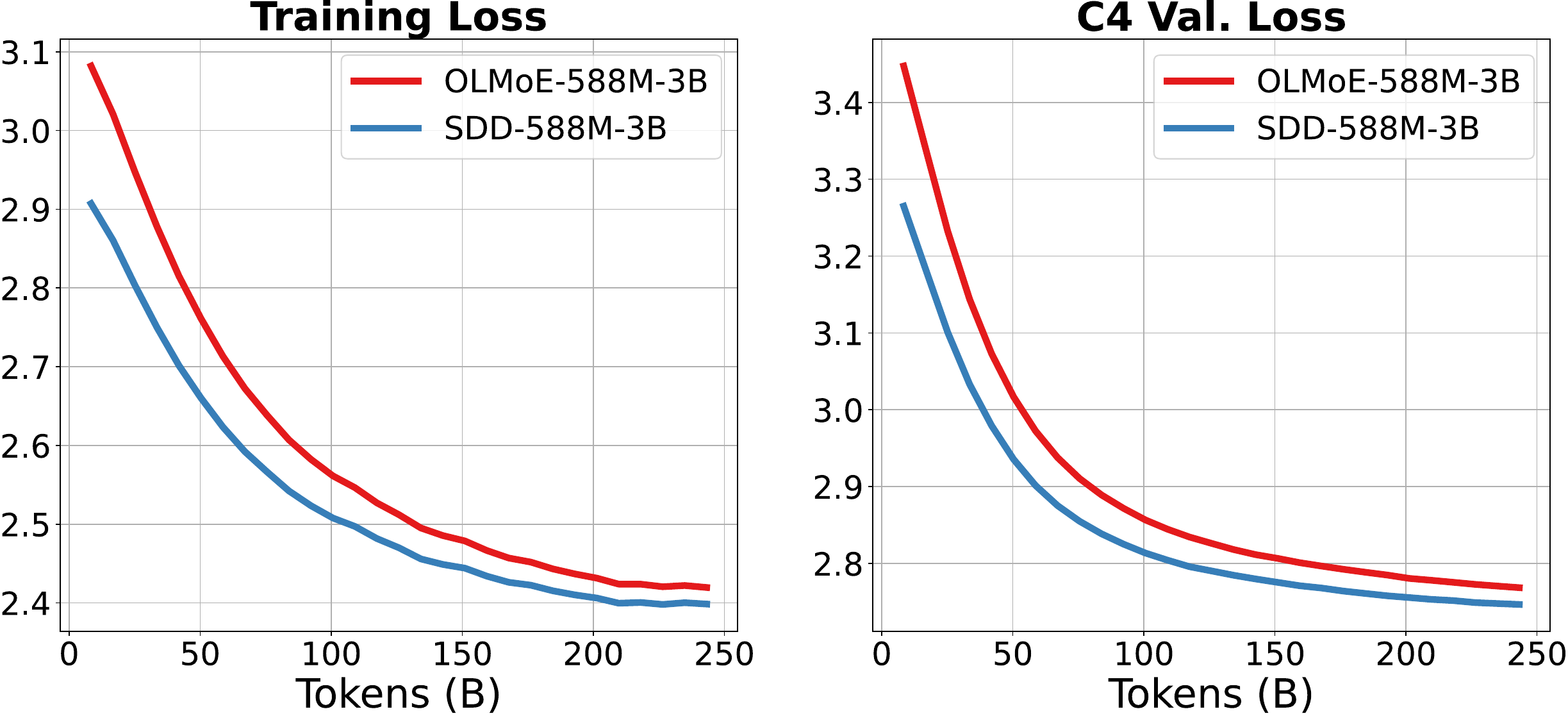}
\vspace{-18pt}
\caption{Training and Validation Loss on C4 for MoE Models with 250 Billion Tokens: Comparison of OLMoE-588M-3B (Pre-Norm) and SDD-588M-3B (Post-Norm).
}
\vspace{-8pt}
\label{fig:overall training results for moe model}
\end{figure}

\textbf{Dowmstream Evaluation.} Table~\ref{table: evaluation results on dense model} and Figure~\ref{fig:overall evaluation results for dense model} summarize downstream results across MMLU, HellaSwag, ARC-Challenge, ARC-Easy, Winogrande, Openbook QA, and COPA. SDD-1B consistently outperforms its counterparts, achieving the highest average accuracy of 54.04\%, surpassing OLMo2-1B (52.01\%), PostNorm-1B (52.03\%), and DeepNorm-1B (51.00\%). Notable gains include a 3.46\% and 2.67\% improvement over the second-best model on ARC-Challenge (37.57\%) and HellaSwag (59.65\%), respectively. These results reinforce SDD-1B's effectiveness in capturing complex linguistic patterns and improving generalization across diverse benchmarks.

\subsection{Results on MoE Model}
We evaluate SDD on OLMoE-588M-3B, an MoE model with 588M active parameters out of 3.4B total \citep{muennighoff2024olmoeopenmixtureofexpertslanguage}. Due to computational constraints, we compare it to the baseline OLMoE-588M-3B with identical hyperparameters. SDD introduces only a 0.1\% increase in parameters due to the learnable scaling vector $\alpha$, ensuring a fair comparison without modifying training settings.

\begin{figure}[!ht]
\centering
\includegraphics[width=\linewidth]{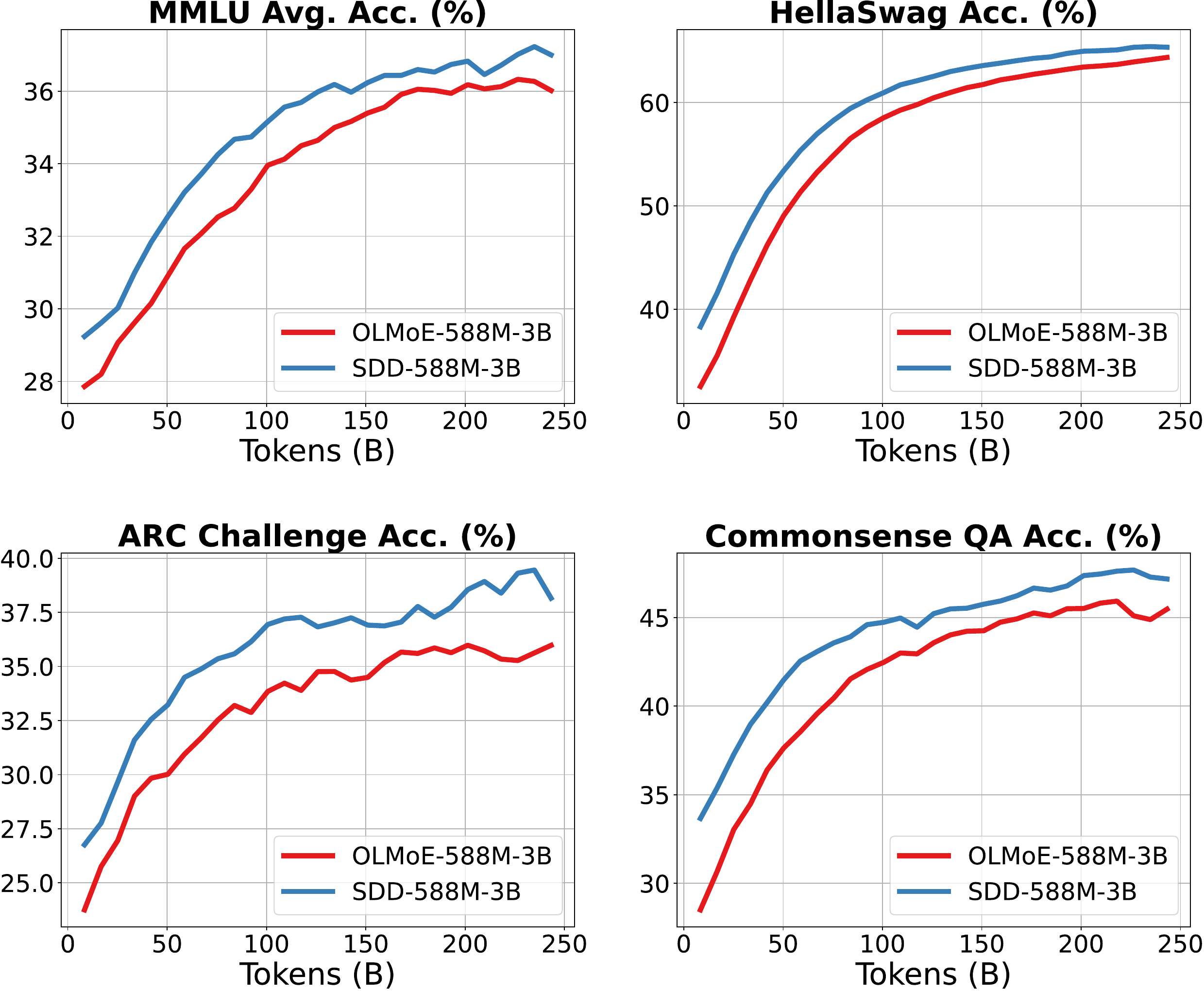}
\vspace{-18pt}
\caption{Downstream performance on MMLU, HellaSwag, ARC-Challenge, and Commonsense for MoE models with 250 billion training tokens.
}
\vspace{-16pt}
\label{fig:overall evaluation results for moe model}
\end{figure}

\textbf{Training dynamics of MoE model.}
Figure~\ref{fig:overall training results for moe model} presents the training and validation loss curves for MoE models trained on 250B tokens. SDD-588M-3B consistently achieves lower losses than OLMoE-588M-3B, demonstrating improved convergence and stability. This suggests that SDD not only accelerates training but also mitigates optimization challenges common in large-scale MoE models.

\textbf{Dowmstream Evaluation.}
Figure~\ref{fig:overall evaluation results for moe model} shows that SDD-588M-3B outperforms OLMoE-588M-3B across all benchmarks, particularly in MMLU, which evaluates multi-domain reasoning. More metrics are available in Appendix~\ref{sec:additional results on moe model}. These improvements underscore SDD's capacity to enhance generalization and capture intricate linguistic patterns. Overall, SDD boosts both training efficiency and downstream performance in MoE architectures, providing a robust and scalable solution for large-scale model optimization.

\subsection{Ablation Study}
\textbf{Gradient Visualization.}
\begin{figure}[!ht]
\centering
\includegraphics[width=\linewidth]{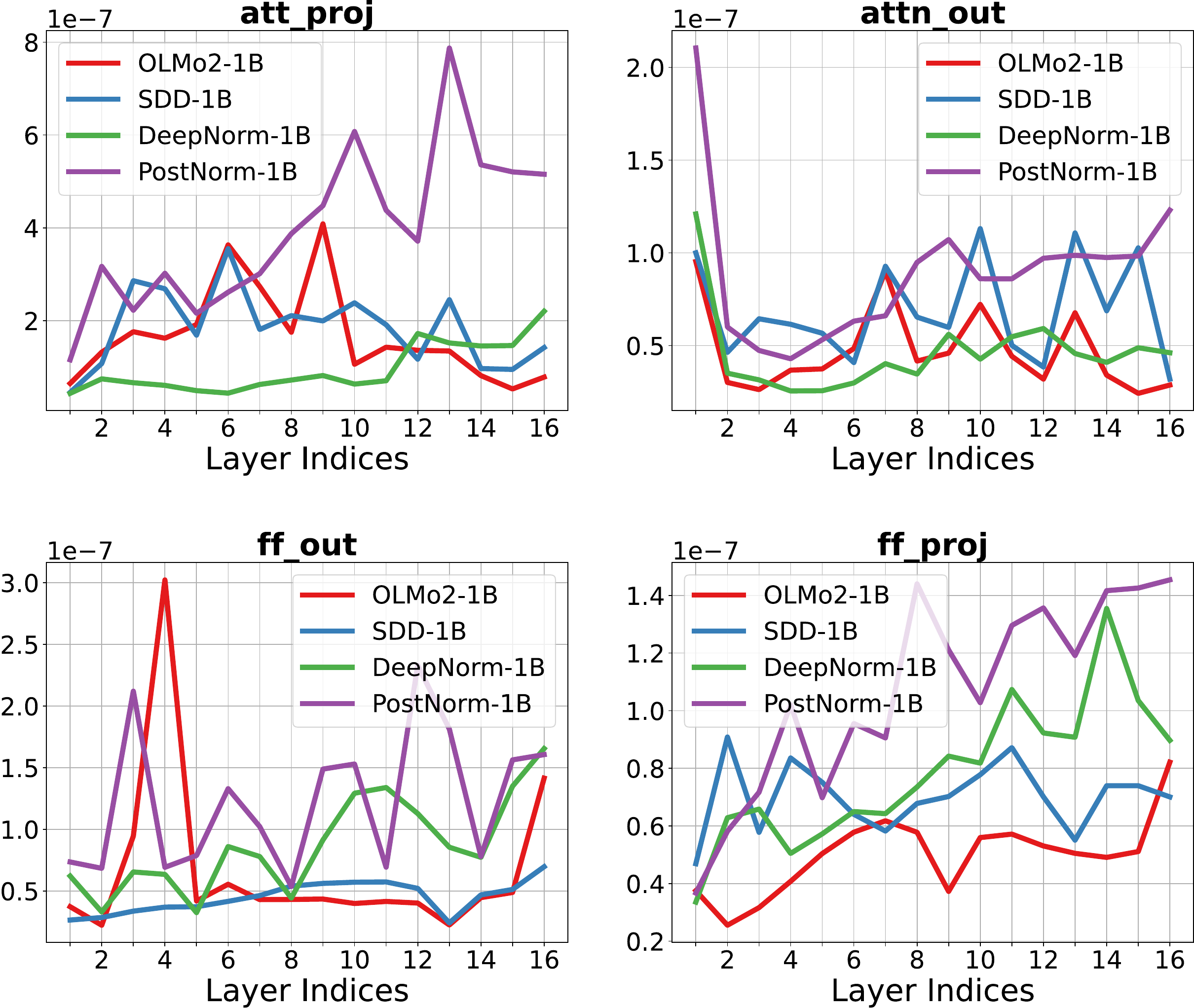}
\vspace{-18pt}
\caption{Comparison of Gradient Norms Across Layers. We compare four methods: OLMo2-1B (Pre-Norm), PostNorm-1B, DeepNorm-1B, and SDD-1B (all Post-Norm). ``att\_proj'' refers to the query/key/value projection, ``attn\_out'' to the attention output projection, ``ff\_proj'' to the gating and first FC layer in the feed-forward network (FFN), and ``ff\_out'' to the second FC layer in the FFN. SDD-1B demonstrates notably stable gradient norms, effectively addressing gradient explosion and vanishing.}
\vspace{-12pt}
\label{fig:gradient analysis}
\end{figure}
Figure~\ref{fig:gradient analysis} compares gradient norms across layers for OLMo2-1B (Pre-Norm), PostNorm-1B, DeepNorm-1B (both Post-Norm), and SDD-1B (Post-Norm). SDD-1B maintains significantly more stable gradient norms, mitigating gradient explosion and vanishing, which commonly affect Post-Norm variants. This stability improves optimization and training robustness, especially in deep networks, making SDD particularly effective for large-scale models.

\begin{figure}[!ht]
\centering
\includegraphics[width=\linewidth]{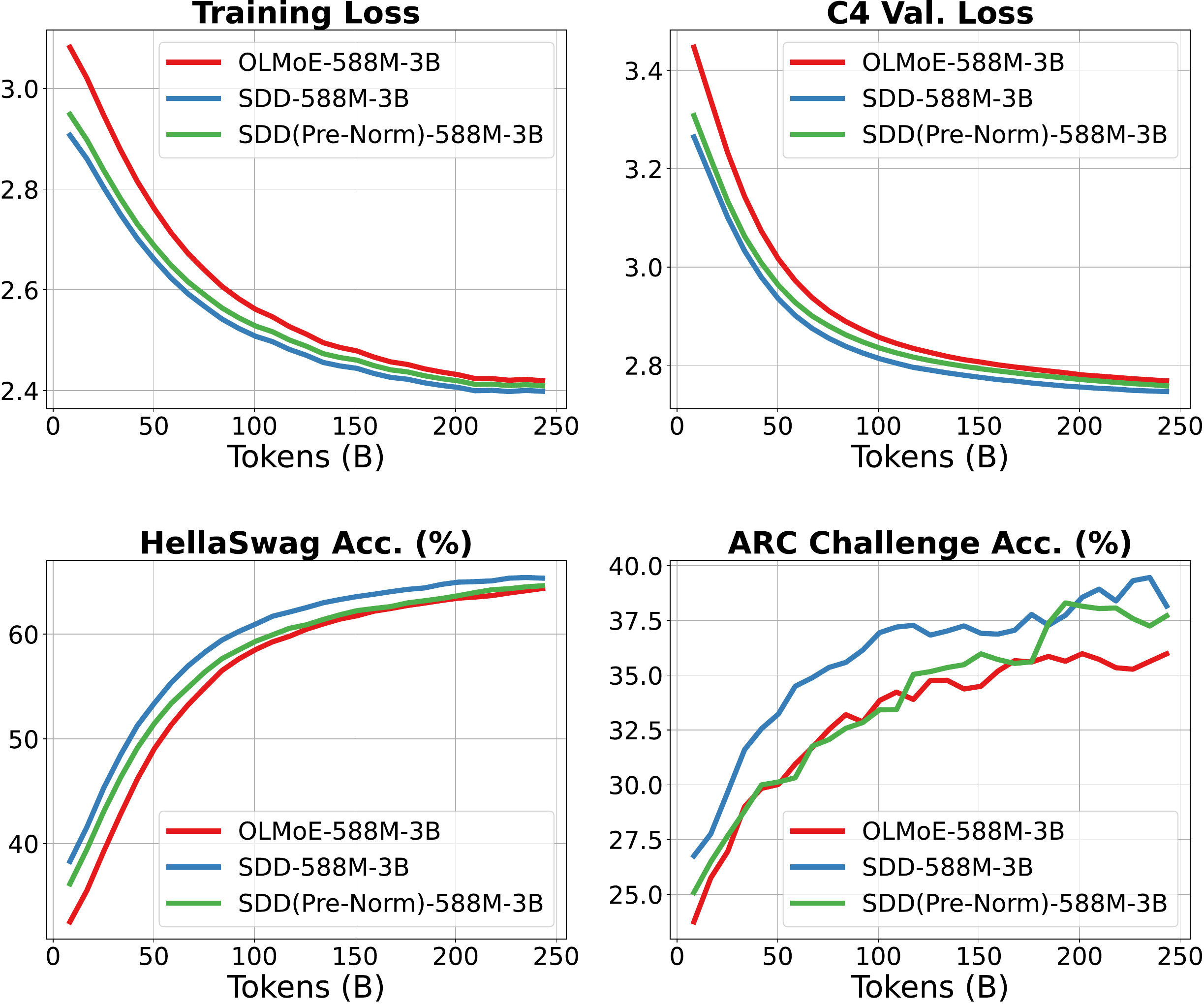}
\vspace{-0.7cm}
\caption{Training and downstream performance of SDD-588M-3B with Pre-Norm and Post-Norm compared to OLMoE-588M-3B (Pre-Norm). Models trained on 250 billion tokens show that SDD improves convergence speed and downstream accuracy in the Pre-Norm setting. Switching to Post-Norm with SDD yields even greater performance gains.}
\label{fig:Pre-Norm}
\end{figure}

\textbf{SDD on Pre-Norm.}
Figure~\ref{fig:Pre-Norm} evaluates SDD-588M-3B under both Pre-Norm and Post-Norm settings. When applied to Pre-Norm, SDD accelerates convergence and enhances downstream accuracy. Further gains are observed when transitioning from Pre-Norm to Post-Norm, highlighting SDD's adaptability and effectiveness in improving training stability and generalization.

\begin{figure}[!t]
\centering
\begin{minipage}{0.49\linewidth}
    \centering
    \includegraphics[width=\linewidth]{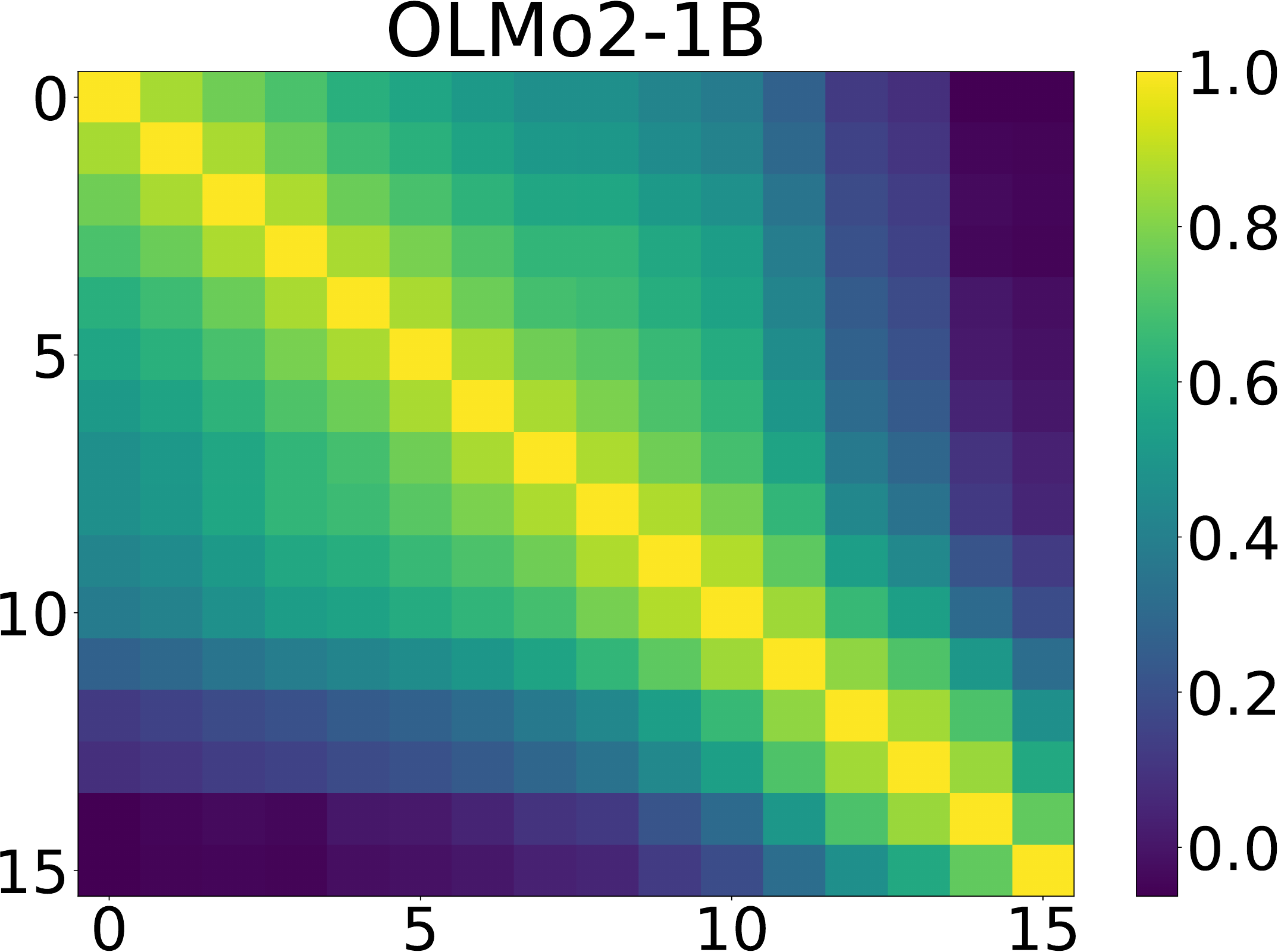}
    \vspace{-0.3cm}
    \label{fig:olmo2_sim}
\end{minipage}
\hfill
\begin{minipage}{0.49\linewidth}
    \centering
    \includegraphics[width=\linewidth]{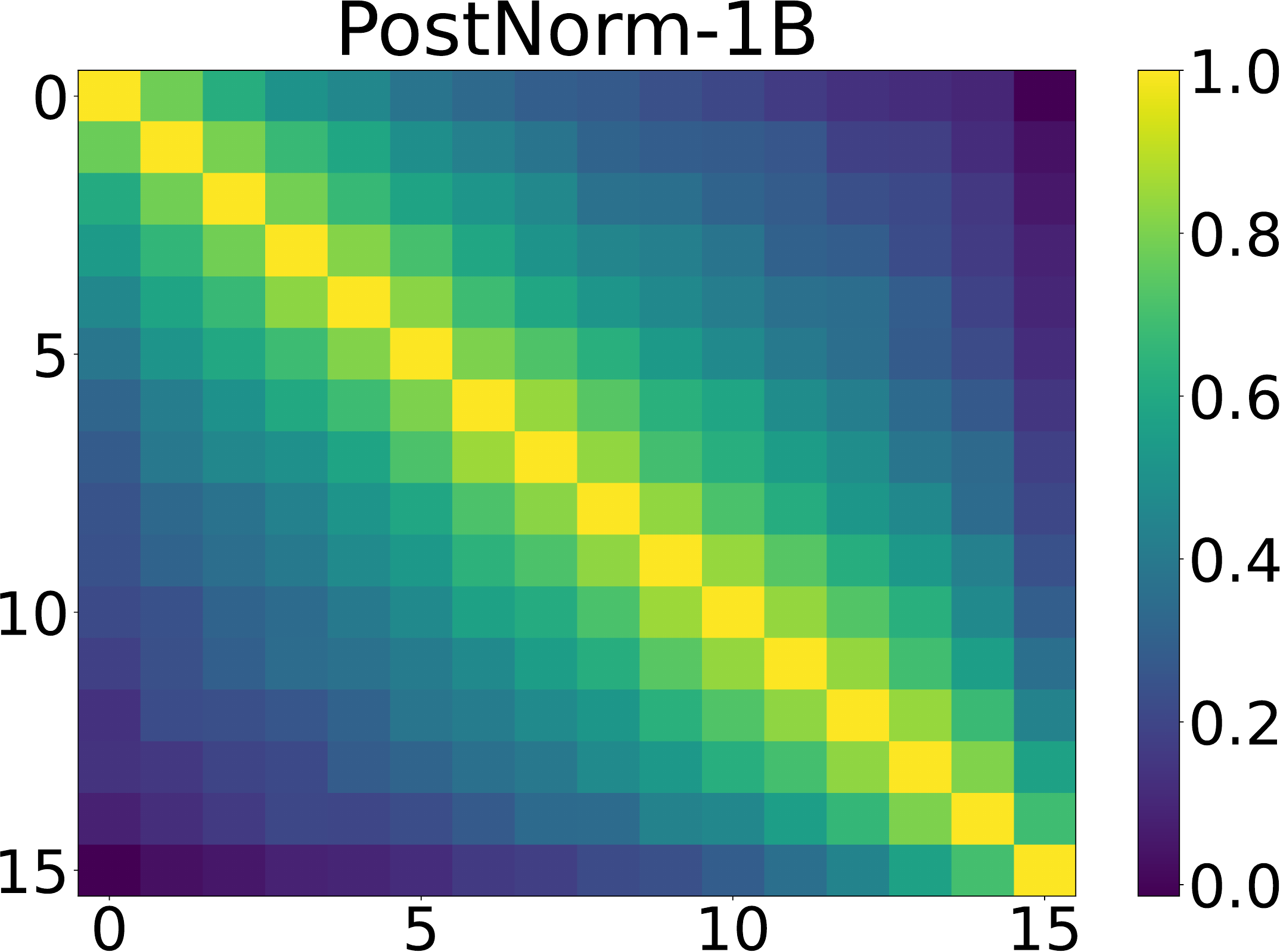}
    \vspace{-0.3cm}
    \label{fig:postnorm_sim}
\end{minipage}

\begin{minipage}{0.49\linewidth}
    \centering
    \includegraphics[width=\linewidth]{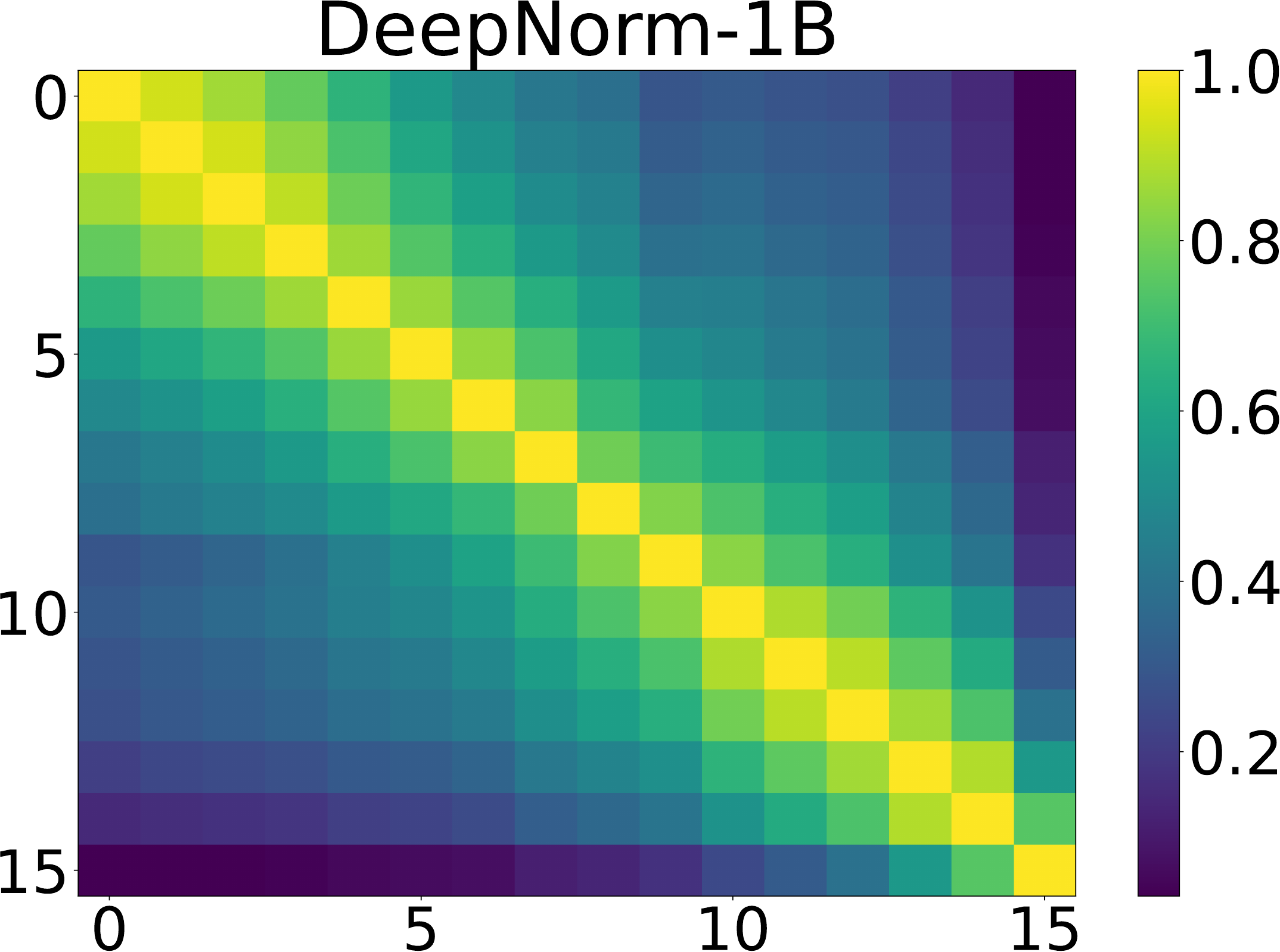}
    \vspace{-0.3cm}
    \label{fig:deepnorm_sim}
\end{minipage}
\hfill
\begin{minipage}{0.49\linewidth}
    \centering
    \includegraphics[width=\linewidth]{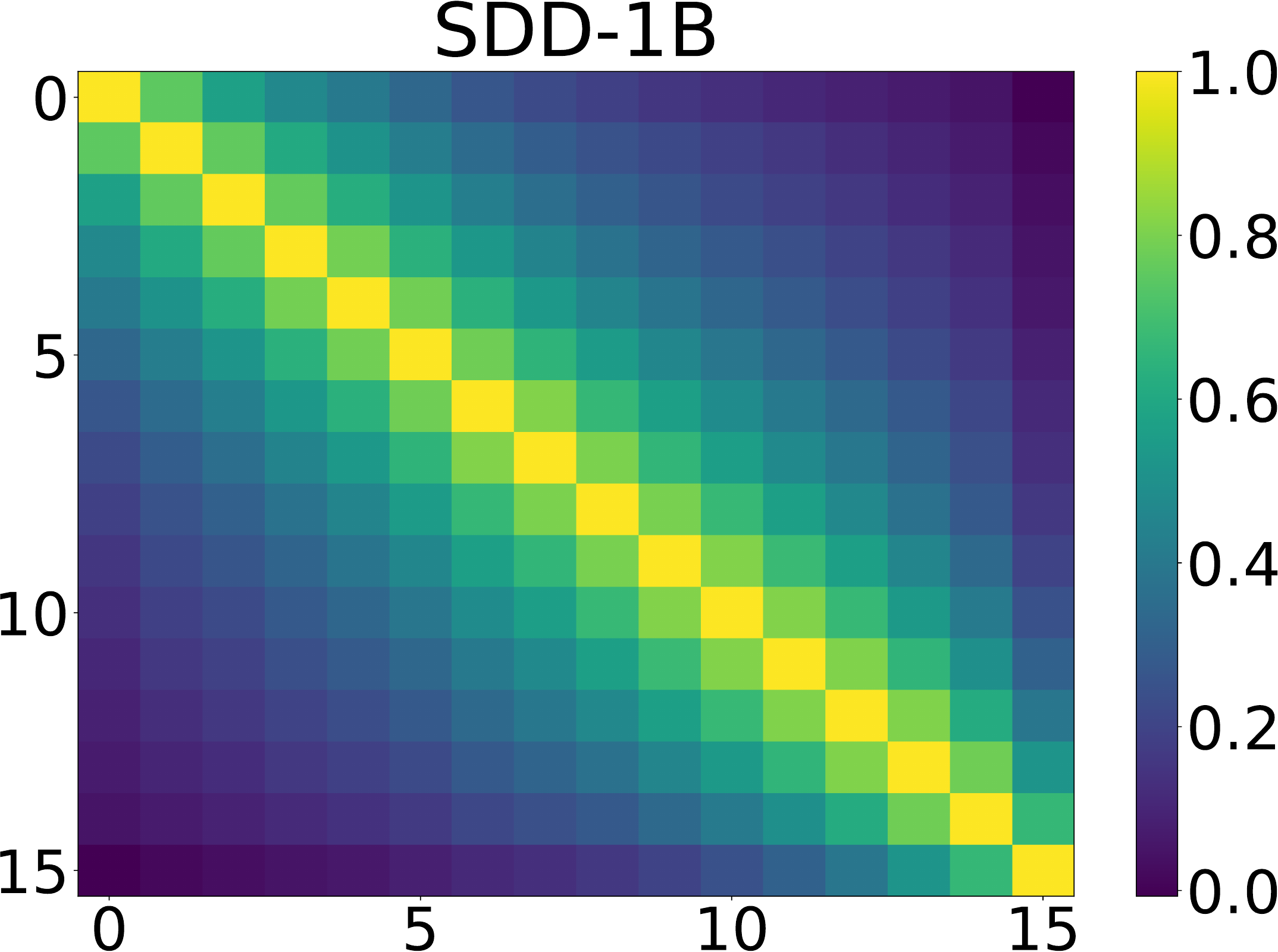}
    \vspace{-0.3cm}
    \label{fig:sdd_sim}
\end{minipage}
\vspace{-8pt}
\caption{Layer-Wise Feature Similarity Across Normalization Methods. This figure compares feature similarity across layers in OLMo2-1B, PostNorm-1B, DeepNorm-1B, and SDD-1B. SDD-1B achieves the highest inter-layer similarity, indicating more stable feature propagation.}
\label{fig:similarity_comparison}
\vspace{-12pt}
\end{figure}

\textbf{Layer-wise Similarity.} Figure~\ref{fig:similarity_comparison} illustrates inter-layer feature similarity across normalization methods. SDD-1B exhibits the lowest similarity, indicating reduced feature redundancy and effectively mitigating feature collapse. This suggests that SDD promotes more diverse representations across layers, contributing to better optimization and enhanced generalization.

\begin{table}[ht]
    \setlength{\textwidth}{0pt}
\setlength{\tabcolsep}{1.8pt}
\caption{Impact of Hyperparameter Perturbations on Model Performance. ``$-$'' indicates non-convergence. All models are trained on 200B tokens. ``lr$*5$'' refers to a 5x increase in learning rate, ``Initstd$*0.1$'' scales the initialization standard deviation by 0.1, and ``wo Warmup'' denotes the removal of the warmup phase.}
\vspace{-14pt}
\label{table:model hyperparameters}
\begin{center}
\begin{tabular}{l|ccccc}
\toprule
\textbf{Model}  & \textbf{lr}$\mathbf{*5}$ &\textbf{Initstd}$\mathbf{*0.1}$  & \textbf{wo WarmUp}   &\textbf{Loss $\downarrow$} \\
\midrule
OLMo2-581M& \ding{55}  &\ding{55}   &\ding{55}  & 2.85        \\
OLMo2-581M& \ding{51}  &\ding{55}   &\ding{55}  & 2.84        \\
OLMo2-581M& \ding{55}  &\ding{51}   &\ding{55}  & 2.86        \\
OLMo2-581M& \ding{55}  &\ding{55}   &\ding{51}  &  2.85       \\
\midrule
PostNorm-581M& \ding{55}  &\ding{55}   &\ding{55}  & 2.84        \\
PostNorm-581M& \ding{51}  &\ding{55}   &\ding{55}  &   $-$      \\
PostNorm-581M& \ding{55}  &\ding{51}   &\ding{55}  &   $-$      \\
PostNorm-581M& \ding{55}  &\ding{55}   &\ding{51}  &  $-$       \\
\midrule
DeepNorm-581M& \ding{55}  &\ding{55}   &\ding{55}  &  2.84       \\
DeepNorm-581M& \ding{51}  &\ding{55}   &\ding{55}  & $-$        \\
DeepNorm-581M& \ding{55}  &\ding{51}   &\ding{55}  &  $-$       \\
DeepNorm-581M& \ding{55}  &\ding{55}   &\ding{51}  & 2.87        \\
\midrule
SDD-581M& \ding{55}  &\ding{55}   &\ding{55}  &2.83         \\
SDD-581M& \ding{51}  &\ding{55}   &\ding{55}  & \textbf{2.81}        \\
SDD-581M& \ding{55}  &\ding{51}   &\ding{55}  & 2.82        \\
SDD-581M& \ding{55}  &\ding{55}   &\ding{51}  & 2.83        \\
\bottomrule
\end{tabular}
\vspace{-10pt}
\end{center}
\end{table}
    
\textbf{Robustness on Hyperparameter Perturbations.} Table~\ref{table:model hyperparameters} assesses model robustness under hyperparameter variations, including increased learning rates, reduced initialization scale, and removal of warmup. While PostNorm-581M and DeepNorm-581M fail to converge under certain conditions, SDD-581M consistently stabilizes training and achieves lower loss, demonstrating resilience to hyperparameter changes.

\begin{figure}[!ht]
\centering
\includegraphics[width=\linewidth]{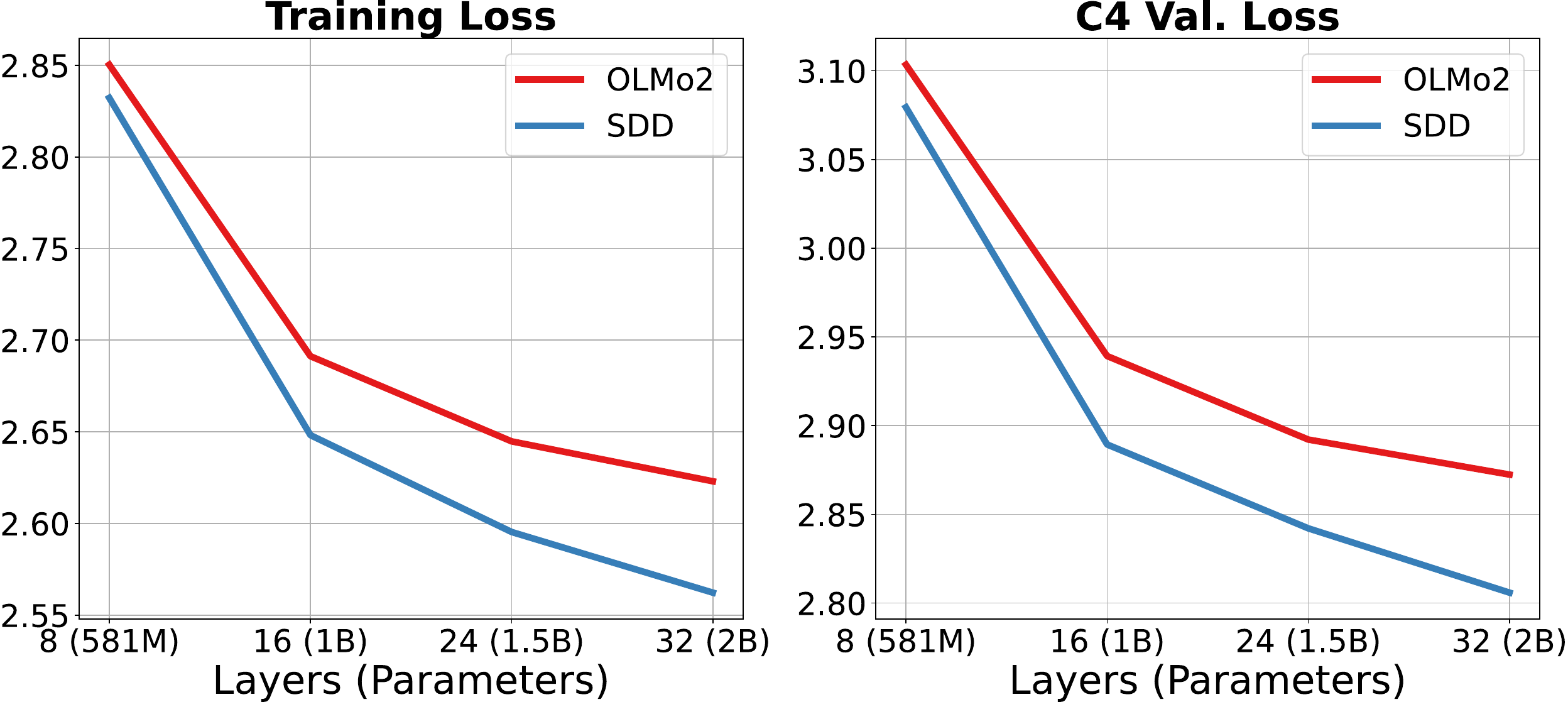}
\vspace{-18pt}
\caption{Scaling with model depth: OLMo2-1B (Pre-Norm) vs. SDD-1B (Post-Norm). All models are trained on 200 billion tokens, with only the number of layers varied. SDD shows superior scaling behavior as model depth increases, highlighting its robustness in deeper networks.}
\vspace{-12pt}
\label{fig:scaling}
\end{figure}

\textbf{Scaling law for model depth.} Figure~\ref{fig:scaling} compares OLMo2-1B (Pre-Norm) and SDD-1B (Post-Norm) across varying depths. SDD enables deeper models to scale effectively, overcoming training instability that typically limits Post-Norm architectures. This is particularly evident as the depth increases, where SDD maintains stability and ensures smooth optimization. These results further validate SDD's ability to improve convergence and performance in large-scale Transformer models, making it a promising solution for very deep architectures.

\section{Related Work} \label{sec:related work}
\textbf{\noindent{Normalization Techniques in Transformers.}}\
Normalization is essential for stabilizing deep Transformer training \citep{wang2024world, wang2022anchor}, with Layer Normalization (LN) \citep{ba2016layer,wang2022anchor} being the standard. Pre-Norm \citep{xiong2020layer} improves stability but often reduces expressivity, while Post-Norm \citep{vaswani2017attention} enhances generative performance but is prone to gradient explosion in deep networks. Approaches like DeepNorm \citep{wang2024deepnet} and Sandwich-LN \citep{ding2021cogview} aim to address these challenges by balancing stability and expressivity. Our method, Scale-Distribution Decoupling (SDD), builds on these efforts by explicitly disentangling the scale and distribution of the weight matrix, preserving stability while enhancing expressivity and optimizing training.

\textbf{\noindent{Mixture of Experts and Large-Scale Model Training.}}\
The adoption of Mixture of Experts (MoE) architectures \citep{shazeer2017outrageously,fedus2022switch} has allowed for more efficient computation by activating subsets of parameters per forward pass. However, MoE introduces instability in expert selection and training divergence. OLMoE \citep{muennighoff2024olmoeopenmixtureofexpertslanguage} and architectures like Switch Transformers \citep{fedus2022switch} mitigate these issues with improved routing and load balancing. SDD complements these approaches by enhancing convergence and robustness, ensuring MoE models remain stable even under varying hyperparameter settings.

\textbf{\noindent{Scaling and Stability in Large Language Models.}}\
Training stability becomes more difficult as Transformer depth increases, with gradient-related issues like vanishing or exploding gradients. Techniques such as T-Fixup \citep{huang2020improving} and GradNorm \citep{chen2018gradnorm} focus on balancing gradient magnitudes, while Megatron-Init \citep{shoeybi2019megatron} improves initialization. However, these methods primarily address stability from a weight-scaling perspective, rather than tackling optimization dynamics directly. SDD addresses these challenges by improving depth scalability and maintaining stable feature representations across layers, reducing redundancy, and mitigating feature collapse. These advantages make SDD a robust solution for training large-scale Transformers.

By addressing both stability and expressivity, SDD offers a scalable and efficient solution that enhances training stability while preserving the model's capacity to capture complex patterns. This decoupling of scale and distribution ensures robust optimization, enabling effective training of modern Transformer architectures, even in deep or high-dimensional networks, while maintaining model performance.

\section{Conclusion} \label{sec:conclusion}
We propose Scale-Distribution Decoupling (SDD), a method that stabilizes Transformer training by explicitly separating the scale and distribution of fully connected layer parameters. Our theoretical analysis establishes its expressivity and training benefits, while gradient analysis confirms improved stability, reducing the risk of gradient explosion or vanishing. Extensive experiments on both dense and Mixture of Experts (MoE) models demonstrate that SDD accelerates convergence, improves generalization, and enhances robustness to hyperparameter perturbations. Additionally, SDD exhibits superior scalability with depth and fosters more consistent inter-layer representations. By addressing key training challenges, SDD provides a principled approach for improving the efficiency and stability of large-scale language models.

\section*{Impact Statement}
This paper presents work whose goal is to advance the field of 
Machine Learning. There are many potential societal consequences of our work, none which we feel must be specifically highlighted here.

\nocite{langley00}

\bibliography{example_paper.bib}
\bibliographystyle{icml2025}

\newpage
\appendix
\onecolumn

\section{Omitted Proof} \label{appendix:proof of theorem equ}
\begin{proof}
\textit{\noindent{(1) $y = W x \implies y = \alpha \odot \mathrm{norm}(V x)$.}}

Let  $W \in \mathbb{R}^{n\times n}$  be the weight matrix of a fully-connected layer, where each element of  $W$  is sampled from an independent Gaussian distribution $\mathcal{N}(0, \sigma^2 / n)$. Using singular value decomposition (SVD),  $W$  can be written as:
\begin{equation}
    W = U \Sigma V'^\top,
\end{equation}
where  $U \in \mathbb{R}^{n\times n}$  and  $V' \in \mathbb{R}^{n \times n}$  are orthogonal matrices, and  $\Sigma \in \mathbb{R}^{n \times n}$  is a diagonal matrix containing the singular values $\sigma_1, \sigma_2, \dots, \sigma_n$ of  $W$. Substituting  $W$  into  $y = W x$, we can rewrite the output as:
\begin{equation}
y = W x = U \Sigma V'^\top x.
\end{equation}
Let  $z = V'^\top x$. Since  $x \sim \mathcal{N}(0, I)$, the orthogonal transformation $V'^\top x$  preserves the Gaussian distribution of  $x$, meaning  $z \sim \mathcal{N}(0, I)$. According to Theorem 3.1.1 \citep{vershynin2018high},  $\|x\|$ is approximately equal to 1. So for simplicity, we set $\|z\|=1$. The term  $\Sigma z$  scales the components of  $z$  along the singular directions, where:
\begin{equation}
    \Sigma z = [\sigma_1 z_1, \sigma_2 z_2, \dots, \sigma_n z_n]^\top,
\end{equation}
The orthogonal matrix  $U$  then rotates the scaled vector  $\Sigma z$:
\begin{equation}
    y = U \Sigma z.
\end{equation}
Next, we normalize  $y$, effectively removing the rotational effect of $U$:
\begin{equation}
\mathrm{norm}(y) = \mathrm{norm}(U \Sigma z) = \frac{U \Sigma z}{\|U \Sigma z\|} = \frac{U \Sigma z}{\|\Sigma z\|}= U \cdot \mathrm{norm}(\Sigma z).
\end{equation}
where  $\|\Sigma z\|$  denotes the norm of the diagonal matrix $\Sigma z$. Thus, $ U$  can always be absorbed into subsequent layers' mappings in a neural network without affecting the overall output, regardless of whether normalization is applied between  $U$  and the subsequent layers. For example, in Transformer models,  $U$  can propagate through the value, output projection of attention, and feed-forward network (FFN) mappings, making its explicit presence inconsequential. In other words, $y'=\Sigma V'^\top x$ is equivalent in expressiveness to $y=Wx$ in Transformer models. Therefore, we make no distinction between $y'$ and $y$.

After absorbing  $U$, and noting that  $\|z\|=1$  implies  $\mathrm{norm}(z) = z$, the output can be reformulated as:
\begin{equation}
y = \alpha \odot V'^\top x = \alpha \odot \mathrm{norm}(V'^\top x),
\end{equation}
where  $\alpha = \mathrm{diag}(\Sigma) = [\sigma_1, \sigma_2, \dots, \sigma_n]^\top$ captures the scale information of  $W$. Let $V = V'^\top$, then the output  $y = W x $ can be equivalently expressed in the form  $y = \alpha \odot \mathrm{norm}(V x)$.

\textit{\noindent{(2) $y = \alpha \odot \mathrm{norm}(V x) \implies y = W x $.}}

Consider the representation  $y = \alpha \odot \mathrm{norm}(V x)$, where  $\alpha \in \mathbb{R}^n$  is a vector, and  $V \in \mathbb{R}^{n\times n}$  is a general matrix that may not necessarily be orthogonal. To demonstrate that this output can be equivalently expressed as  $y = W x$, the matrix  $V$  is decomposed using singular value decomposition (SVD). Specifically, let:
\begin{equation}
V = P \Lambda Q^\top,
\end{equation}
where  $P \in \mathbb{R}^{n\times n}$  and  $Q \in \mathbb{R}^{n \times n}$  are orthogonal matrices, and  $\Lambda \in \mathbb{R}^{n \times n}$  is a diagonal matrix containing the singular values of  $V$, denoted as $\gamma_1, \gamma_2, \cdots, \gamma_n$.

Substituting the decomposition of  $V$  into the given equation, the output becomes:
\begin{equation}
y = \alpha \odot \mathrm{norm}(V x) = \alpha \odot \mathrm{norm}(P \Lambda Q^\top x).
\end{equation}
Define  $z = Q^\top x$. Since  $x \sim \mathcal{N}(0, I)$, and by Theorem 3.1.1 \citep{vershynin2018high}, $\|x\|$ is approximately 1. For brevity, we assume $\|x\|=1$. The orthogonality of $Q$ guarantees that $\|z\|=1$. Therefore, the expression for  $y$ can now be written as:
\begin{equation}
y = \alpha \odot \mathrm{norm}(P \Lambda z).
\end{equation}
To simplify further, note that the normalization operation satisfies:
\begin{equation}
\mathrm{norm}(P \Lambda z) = P \cdot \mathrm{norm}(\Lambda z).
\end{equation}
For a diagonal matrix  $\Lambda$, the normalization of  $\Lambda z$  can be approximately expressed as:
\begin{equation}
\mathrm{norm}(\Lambda z) = \frac{\Lambda z}{\|\Lambda z\|} \approx \frac{\Lambda z}{\|\Lambda\|}.
 \end{equation}
Here, $\|\Lambda\|=\sqrt{(\gamma_1^2 + \gamma_2^2 + \cdots +\gamma_n^2)/n}$. Substituting this result, the output becomes:
\begin{equation}
y = \alpha \odot P \frac{\Lambda z}{\|\Lambda\|}.
\end{equation}
By substituting back  $z = P^\top x$, we have:
\begin{equation}
y = \alpha \odot P \frac{\Lambda}{\|\Lambda\|} Q^\top x.
\end{equation}
The equivalence to  $y = W x$  is now established by defining:
\begin{equation}
W = \alpha \odot P \frac{\Lambda}{\|\Lambda\|} Q^\top.
\end{equation}
Thus,  $W \in \mathbb{R}^{n\times n}$  is a valid weight matrix that satisfies  $y = W x$  for any  $x$, completing the proof of reverse equivalence.
\end{proof}

\section{Architectural Configuration} \label{appendix: architecture}
Table~\ref{table:model_architecture} presents the architectural specifications of the evaluated models, including the OLMo2-581M, OLMo2-1B OLMo2-1.5B, and OLMo2-2B dense models, as well as the OLMoE-588M-3B Mixture of Experts (MoE) model. Key attributes such as parameter counts, hidden dimensions, attention configurations, and expert routing details are provided for comparison.
\begin{table}[!ht]
\caption{Architectural Configurations of the Dense Model and MoE Model.}
\label{table:model_architecture}
\begin{center}
\begin{tabular}{lccccc}
\toprule
Property               &OLMo2-581M  & OLMo2-1B   & OLMo2-1.5B   & OLMo2-2B   & OLMoE-588M-3B       \\
\midrule
Activate Params       & 581M     & 1B       & 1.5B       & 2B       & 588M            \\
Total Params          & 581M     & 1B       & 1.5B       & 2B       & 3.4B            \\
Hidden Size           & 2048     & 2048     & 2048     & 2048     & 1024            \\
Intermediate Size     &  8192    & 8192     &  8192    & 8192     & 512     \\
GQA Groups            & 8     & 8        & 8     & 8        & 1               \\
Attention Heads       & 32       & 32     & 32       & 32     & 16              \\
Hidden Layers         & 8     & 16       & 24     & 32       & 32              \\
Experts               & $-$      & $-$     & $-$      & $-$     & 64              \\
Topk Experts          & $-$      & $-$     & $-$      & $-$     & 8              \\
Context Length         & 4096    & 4096      & 4096    & 4096      & 4096            \\
Vocabulary Size       & 100278    & 100278     & 100278    & 100278      & 50280           \\
\bottomrule
\end{tabular}
\vspace{-16pt}
\end{center}
\end{table}

\section{Additional Results on Dense models} \label{sec:additional resutls on dense model}
\begin{figure}[!ht]
\centering
\includegraphics[width=0.75\linewidth]{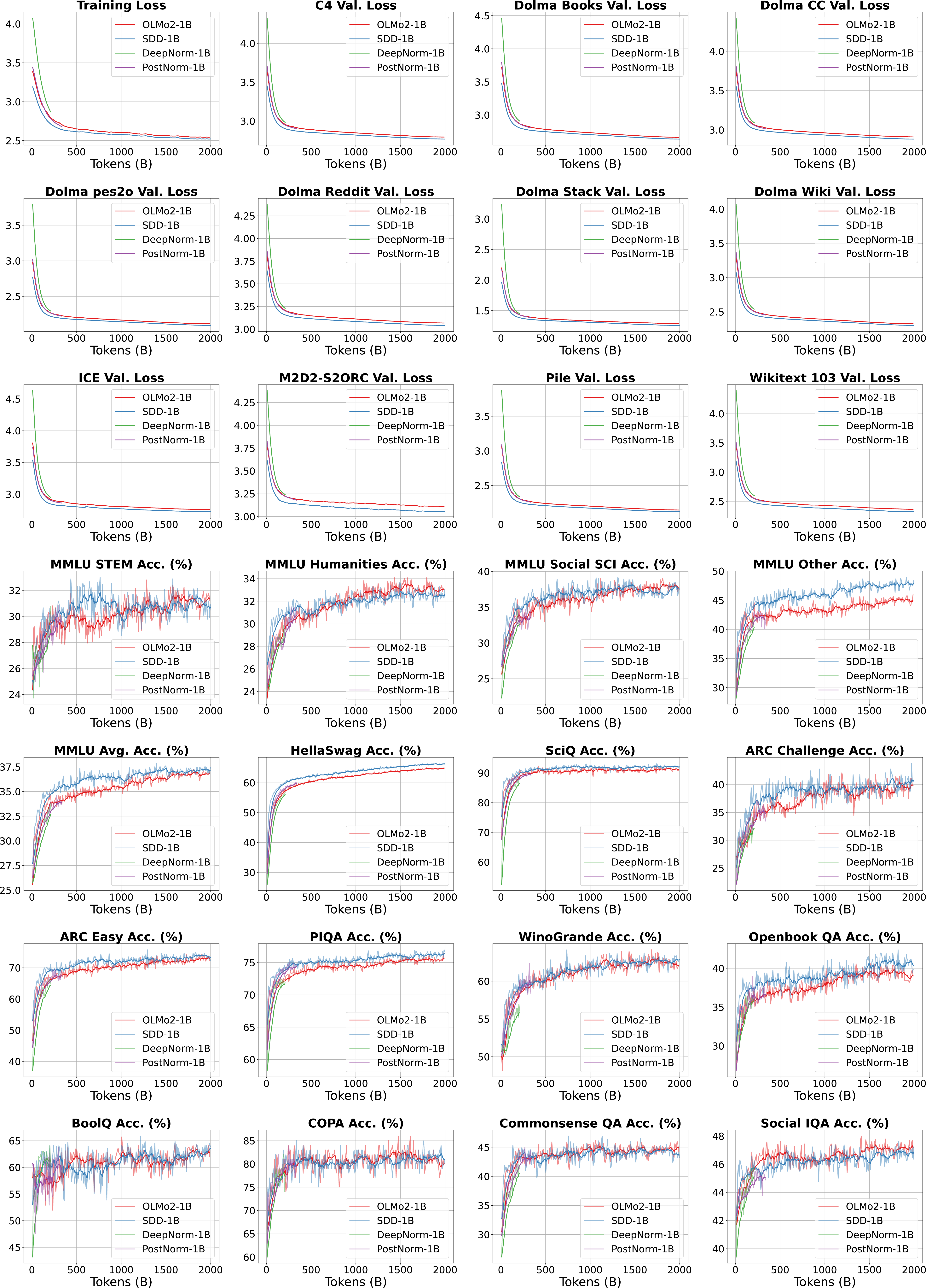}
\caption{Training and Downstream Performance of Dense Models. This figure compares validation loss and downstream task performance for SDD-1B and OLMo2-1B trained on 2T tokens, alongside PostNorm-1B and DeepNorm-1B trained on 200B tokens. SDD-1B exhibits lower loss and superior generalization, demonstrating its effectiveness in large-scale training.}
\label{fig:additional results for dense model}
\vspace{-2pt}
\end{figure}

Figure~\ref{fig:additional results for dense model} presents validation loss and downstream evaluation results for dense models under different training regimes. It compares SDD-1B and OLMo2-1B trained on 2T tokens with PostNorm-1B and DeepNorm-1B trained on 200B tokens. SDD-1B consistently achieves lower validation loss and outperforms all baselines across multiple downstream tasks, highlighting its superior convergence and generalization capabilities. These results further demonstrate the advantages of Scale-Distribution Decoupling (SDD) in stabilizing optimization and improving performance in large-scale language model training.

\section{Additional Results on MoE models} \label{sec:additional results on moe model}
\begin{figure}[th]
\centering
\includegraphics[width=0.75\linewidth]{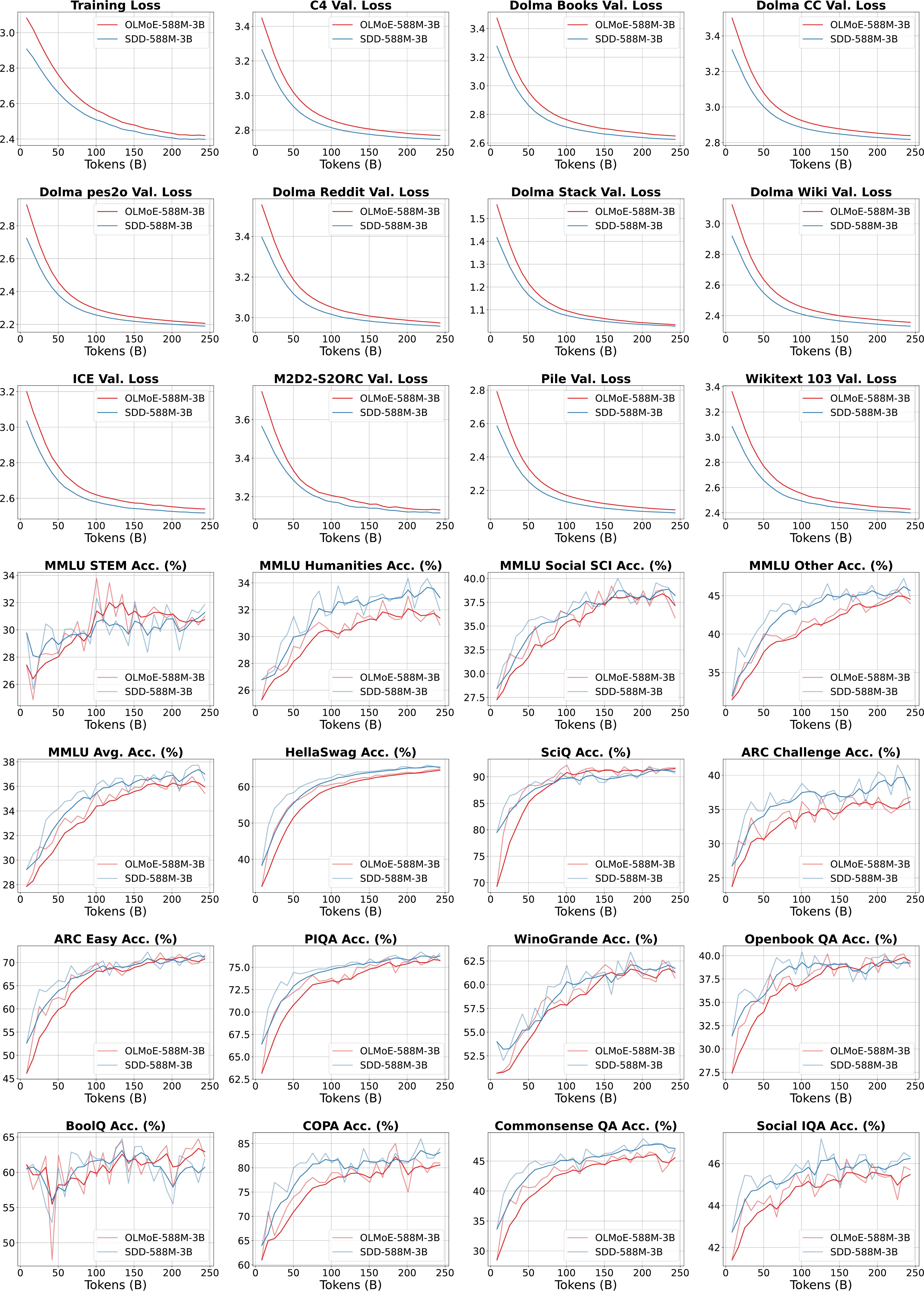}
\caption{Training and Downstream Performance of MoE Models with 250B Tokens. This figure compares the validation loss and downstream task performance of SDD-588M-3B and OLMoE-588M-3B. SDD-588M-3B demonstrates lower loss and superior generalization across benchmarks, highlighting its effectiveness in MoE training.}
\label{fig:additional results for 588M moe model}
\end{figure}

Figure~\ref{fig:additional results for 588M moe model} presents the validation loss and downstream task performance of MoE models trained with 250B tokens, comparing SDD-588M-3B and OLMoE-588M-3B. SDD-588M-3B consistently achieves lower validation loss, indicating improved training stability and efficiency. Additionally, it outperforms OLMoE-588M-3B across multiple benchmarks, demonstrating superior generalization. These results highlight the benefits of Scale-Distribution Decoupling (SDD) in enhancing MoE model optimization, leading to more stable convergence and improved downstream task performance.

\end{document}